\documentclass[letterpaper, 10pt, conference]{ieeeconf}  

\IEEEoverridecommandlockouts                              

\usepackage{epsfig}
\usepackage{graphicx}
\usepackage{amsmath}
\usepackage{amssymb}
\usepackage{gensymb}
\usepackage{subfig}
\usepackage{float}
\usepackage{mathtools}
\usepackage{dblfloatfix}
\usepackage{bm}

\usepackage{comment}
\usepackage{multirow}
\usepackage{array}
\usepackage{booktabs}
\usepackage[export]{adjustbox}
\usepackage{xfrac}
\usepackage{stmaryrd}

\usepackage{comment}
\usepackage{multirow}
\usepackage{array}
\usepackage{tabularx}
\usepackage{booktabs}
\usepackage{xspace} 
\usepackage[ancient]{flushend}

\usepackage[pagebackref=true,breaklinks=true,letterpaper=true,colorlinks,bookmarks=false]{hyperref}

\makeatletter
\let\NAT@parse\undefined
\makeatother

\usepackage[numbers,sort&compress]{natbib}
\bibpunct{[}{]}{,}{n}{}{;}

\def\iccvPaperID{2553} 

\begin{document}

\title{Self-Supervised Camera Self-Calibration from Video}

\author{
Jiading Fang \qquad
Igor Vasiljevic \qquad
Vitor Guizilini \qquad
Rareș Ambruș\\
Greg Shakhnarovich\qquad
Adrien Gaidon\qquad
Matthew R.\ Walter 
\thanks{Jiading Fang, Igor Vasiljevic, Greg Shakhnarovich, and Matthew R.\ Walter are with the Toyota Technological Institute at Chicago {\tt \small\{fjd,ivas,greg,mwalter\}@ttic.edu}}
\thanks{Vitor Guizilini, Rares Ambrus, and Adrien Gaidon are with the Toyota Research Institute {\tt \small \{vitor.guizilini, rares.ambrus, adrien.gaidon\}@tri.global}}}

\maketitle


\begin{abstract}
Camera calibration is integral to robotics and computer vision algorithms that seek to infer geometric properties of the scene from visual input streams. 
In practice, calibration is a laborious procedure
requiring specialized data collection and careful tuning.
This process must be repeated whenever the parameters of the camera change, which can be a frequent occurrence for mobile robots and autonomous vehicles. 
In contrast, self-supervised depth and ego-motion estimation approaches can bypass explicit calibration by inferring per-frame projection models that optimize a view-synthesis objective.
In this paper, we extend this approach to explicitly calibrate a wide range of cameras from raw videos in the wild. We propose a learning algorithm to regress  per-sequence calibration parameters using an efficient family of general camera models.
Our procedure achieves self-calibration results with sub-pixel reprojection error, outperforming other learning-based methods.  We validate our approach on a wide variety of camera geometries, including perspective, fisheye, and catadioptric.  Finally, we show that our approach leads to improvements in the downstream task of depth estimation, achieving state-of-the-art results on the EuRoC dataset with greater computational efficiency than contemporary methods.
The project page: \url{https://sites.google.com/ttic.edu/self-sup-self-calib}
\end{abstract}

\section{Introduction}

Cameras provide rich information about the scene, while being small, lightweight, inexpensive, and power efficient. Despite their wide availability, camera calibration largely remains a manual, time-consuming process that typically requires collecting images of known targets (e.g., checkerboards) as they are deliberately moved in the scene~\cite{zhang2000flexible}. While applicable to a wide range of camera models~\cite{scaramuzza2006flexible,kannala2006generic,grossberg2001general}, this process is tedious and has to be repeated whenever the camera parameters change. A number of methods perform calibration ``in the wild''~\cite{caprile1990using, pollefeys1997stratified, cipolla1999camera}. However, they rely on strong assumptions about the scene structure, which cannot be met during deployment in unstructured environments. Learning-based methods relax these assumptions, and regress camera parameters directly from images, either by using labelled data for supervision~\cite{bogdan2018deepcalib} or by extending the framework of self-supervised depth and ego-motion estimation~\cite{garg2016unsupervised, zhou2017unsupervised} to also learn per-frame camera parameters~\cite{gordon2019depth, vasiljevic2020neural}.

\begin{figure}[t!]
\vspace{-2mm}
\centering
\subfloat[Input]{
\includegraphics[width=0.15\textwidth,height=1.8cm]{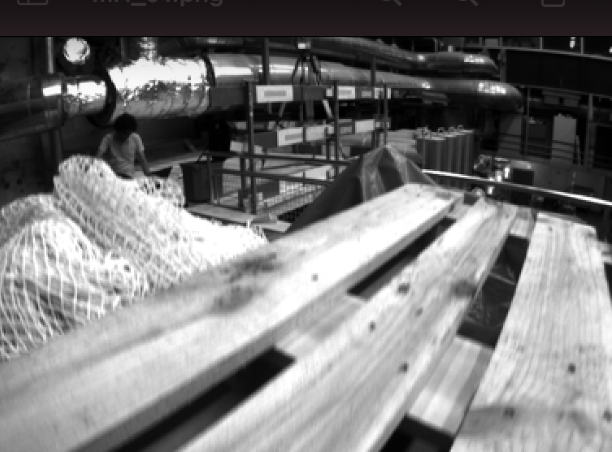}
}
\subfloat[Predicted depth]{
\includegraphics[width=0.15\textwidth,height=1.8cm]{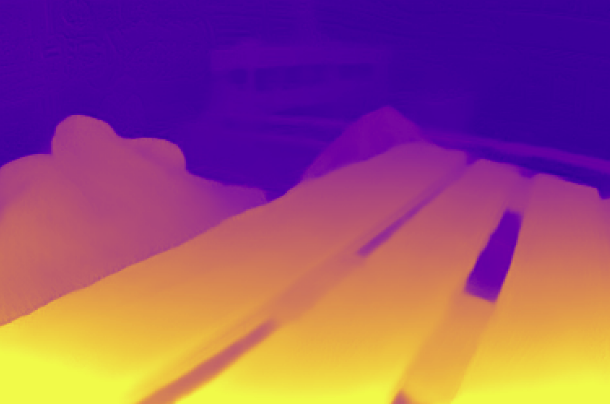}
}
\subfloat[Rectified image]{
\includegraphics[width=0.15\textwidth,height=1.8cm]{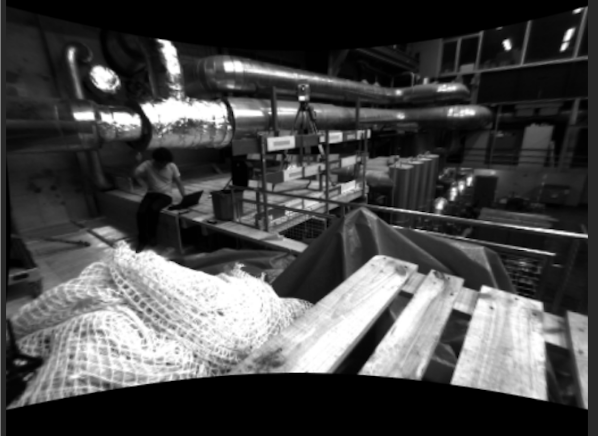}
}
\\
\subfloat[Example re-calibration results from perturbations of a camera parameter.
]{
\includegraphics[width=0.47\textwidth,height=3.4cm]{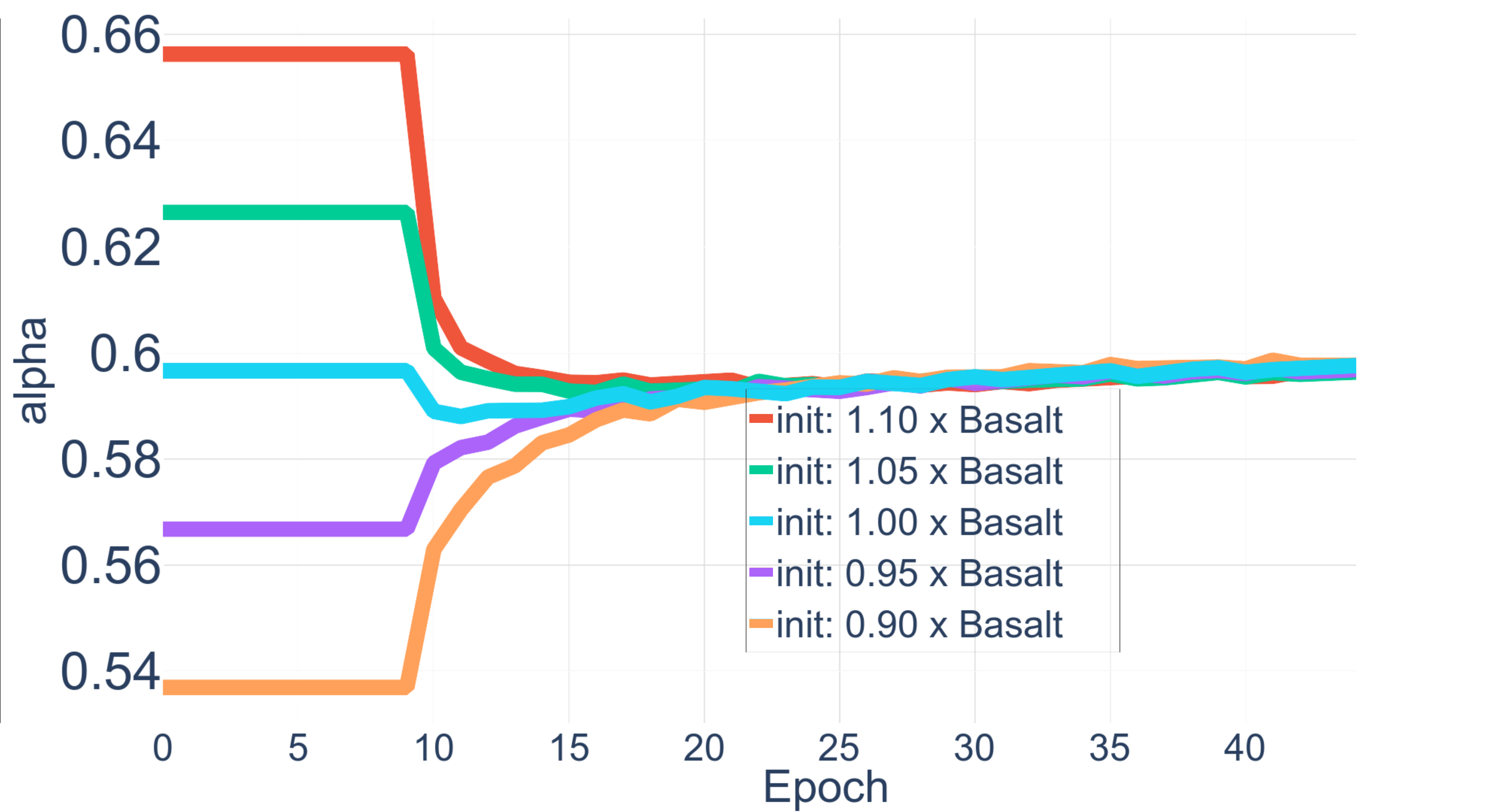}
}

\caption{\textbf{Our self-supervised self-calibration procedure} can recover accurate parameters for a wide range of cameras using a structure-from-motion objective on raw videos (EuRoC dataset, top), enabling on-the-fly re-calibration and robustness to intrinsics perturbation (bottom).}
\label{fig:teaser}
\vspace{-5mm}
\end{figure}

While these methods enable learning accurate depth and ego-motion without calibration, they are either over-parameterized~\cite{vasiljevic2020neural} or limited to near-pinhole cameras~\cite{gordon2019depth}. In contrast, we propose a self-supervised camera calibration algorithm capable of learning expressive models of different camera geometries in a computationally efficient manner. In particular, our approach adopts a family of general camera models~\cite{usenko2018double} that scales to higher resolutions than previously possible, while still being able to model highly complex geometries such as catadioptric lenses. Furthermore, our framework learns camera parameters per-sequence rather than per-frame, resulting in self-calibrations that are more accurate and more stable than those achieved using contemporary learning methods. We evaluate the reprojection error of our approach compared to conventional target-based calibration routines, showing comparable sub-pixel performance despite only using raw videos at training time.

Our contributions can be summarized as follows:
\begin{itemize}
\item We propose to self-calibrate a variety of generic camera models from raw video using self-supervised depth and pose learning as a proxy objective, providing for the first time a calibration evaluation of camera model parameters \textbf{learned purely from self-supervision}.
\item We demonstrate the utility of our framework on challenging and radically different datasets, learning depth and pose on perspective, fisheye, and catadioptric images without architectural changes.
\item We achieve \textbf{state-of-the-art depth evaluation results on the challenging EuRoC MAV dataset} by a large margin, using our proposed self-calibration framework. 
\end{itemize}


    


\section{Related Work}
\textbf{Camera Calibration.}
Traditional calibration for a variety of camera models uses targets such as checkerboards or AprilTags to generate 2D-3D correspondences, which are then used in a bundle adjustment framework to recover relative poses as well as intrinsics~\cite{zhang2000flexible, hartley2000zisserman}. Targetless methods typically make strong assumptions about the scene, such as the existence of vanishing points and known (Manhattan world) scene structure~\cite{caprile1990using, pollefeys1997stratified, cipolla1999camera}.  
While highly accurate, these techniques require a controlled setting and manual target image capture to re-calibrate. Several models are implemented in OpenCV~\cite{bradski2000opencv}, kalibr~\cite{rehder2016extending}.
These methods require specialized settings to work, limiting their generalizability. 

\textbf{Camera Models.}
The pinhole camera model is ubiquitous in robotics and computer vision~\cite{leonard08,urmson2008autonomous} and is especially common in recent deep learning architectures for depth estimation~\cite{zhou2017unsupervised}.  There are two main families of models for high-distortion cameras. The first is the ``high-order polynomial'' distortion family that includes pinhole radial distortion~\cite{fryer1986lens}, omnidirectional~\cite{scaramuzza2006flexible}, and Kannala-Brandt~\cite{kannala2006generic}. The second is the ``unified camera model'' family that includes the Unified Camera Model (UCM)~\cite{geyer2000unifying}, Extended Unified Camera Model (EUCM)~\cite{khomutenko2015enhanced},
and Double Sphere Camera Model (DS)~\cite{usenko2018double}. Both families are able to achieve low reprojection errors for a variety of different camera geometries~\cite{usenko2018double}, however the unprojection operation of the ``high-order polynomial'' models requires solving for the root of a high-order polynomial, typically using iterative optimization, which is a computationally expensive operation. Further, the process of calculating gradients for these models is non-trivial.
In contrast, the ``unified camera model'' family has an easily computed, closed-form unprojection function. While our framework is applicable to high-order polynomial models, we choose to focus on the unified camera model family in this paper.

\textbf{Learning Camera Calibration.}
Work in learning-based camera calibration can be divided into two types: \emph{supervised} approaches that leverage ground-truth calibration parameters or synthetic data to train single-image calibration regressors; and \emph{self-supervised} methods that utilize only image sequences. Our proposed method falls in the latter category, and aims to self-calibrate a camera system using only image sequences.
Early work on applying CNNs to camera calibration focused on regressing the focal length~\cite{workman2015deepfocal} or horizon lines~\cite{workman2016horizon}; synthetic data was used for distortion calibration~\cite{rong2016radial}  and fisheye rectification~\cite{yin2018fisheyerecnet}.  Using panorama data to generate images with a wide variety of intrinsics, \citet{lopez2019deep} are able to estimate both extrinsics (tilt and roll) and intrinsics (focal length and radial distortion).  DeepCalib~\cite{bogdan2018deepcalib} takes a similar approach:  given a panoramic dataset, generate projections with different focal lengths. Then, they train a CNN to regress from a set of synthetic images $I$ to their (known) focal lengths $f$. Typically, training images are generated by taking crops of the desired focal lengths from $360$ degree panoramas~\cite{hold2018perceptual, zhu2020single}. While this can be done for any kind of image, and does not require image sequences, it does require access to panoramic images. Furthermore, the warped ``synthetic'' images are not the true 3D-2D projections. This approach has been extended to pan-tilt-zoom~\cite{zhang2020deepptz} and fisheye~\cite{yin2018fisheyerecnet} cameras.
Methods also exist for specialized problems like undistorting portraits~\cite{zhao2019learning}, monocular 3D reconstruction~\cite{yin2021learning}, and rectification~\cite{yang2021progressively, liao2021deep}.

\textbf{Self-Supervised depth and ego-motion}. Self-supervised learning has also been used to learn camera parameters from geometric priors.  Gordon et al.~\cite{gordon2019depth} learn a pinhole and radial distortion model, while Vasiljevic et al.~\cite{vasiljevic2020neural} learn a generalized central camera model applicable to a wider range of camera types, including catadioptric. These methods both learn calibration on a per-frame basis, and do not offer a calibration evaluation of their learned camera model.  Furthermore, while \citet{vasiljevic2020neural} is much more general than \citet{gordon2019depth}, it is limited to fairly low resolutions by the complex and approximate generalized projection operation. In our work, we trade some degree of generality (i.e., a global, central vs.\ per-pixel model) for a closed-form and efficient projection operation and ease of calibration evaluation.


\section{Methodology}
First, we describe the self-supervised monocular depth learning framework that we use as proxy for self-calibration. Then we describe the family of unified camera models we consider and how we learn their parameters end-to-end. 

\subsection{Self-Supervised Monocular Depth Estimation}
\label{subsec:monodepth}
\begin{figure}[!t]
    \centering
    \includegraphics[width=\linewidth]{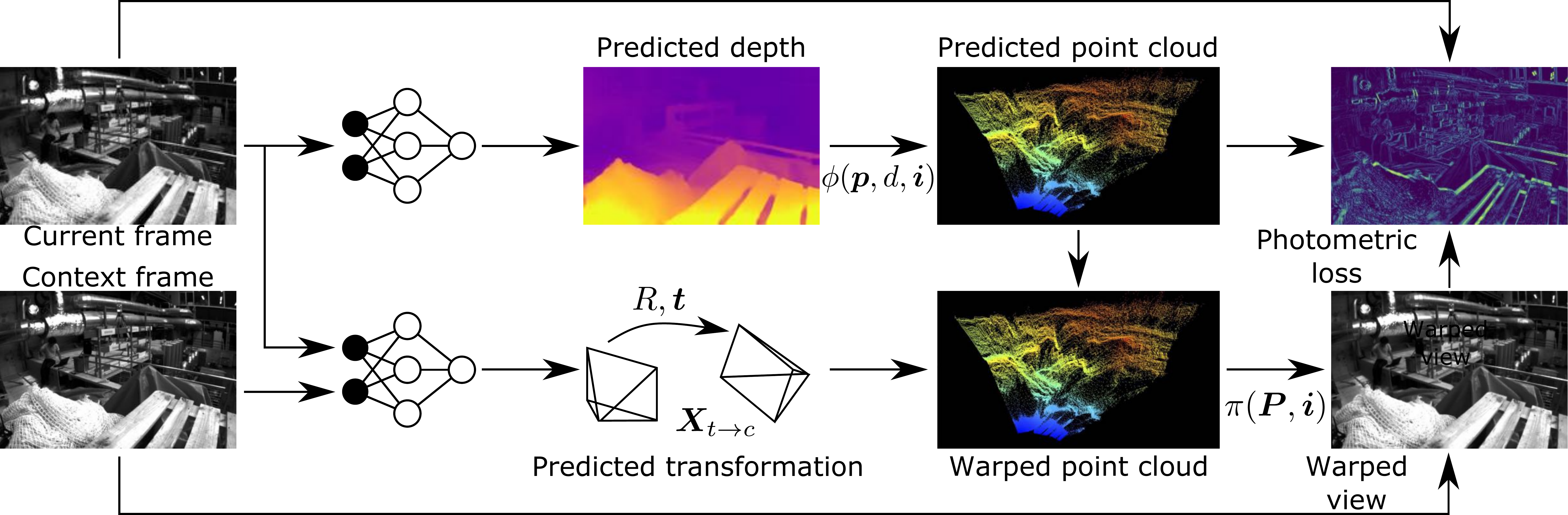}
    \caption{\textbf{Our self-supervised self-calibration architecture.}  We use gradients from the photometric loss to update the parameters of a unified camera model (Fig.\ \ref{fig:ucm_figure}).
    }
    \label{fig:ssl}
\vspace*{-2mm}
\end{figure}

Self-supervised depth and ego-motion architectures consist of a depth network that produces depth maps $\hat{D}_{t}$ for a target image $I_t$, as well as a pose network that predicts the relative rigid-body transformation between target $t$ and context $c$ frames, $\hat{\bm{X}}^{t \to c} = \begin{psmallmatrix}{\hat{\bm{R}}^{t\to c}} & {\hat{\bm{t}}^{t\to c}}\\ \bm{0} & \bm{1}\end{psmallmatrix} \in \text{SE(3)}$.
We train the networks jointly by minimizing the photometric reprojection error between the actual target image $I_t$ and a synthesized image $\hat{I}_t$ generated by projecting pixels from the context image $I_c$ (usually preceding or following $I_t$ in a sequence) onto the target image $I_t$ using the predicted depth map $\hat{D}_{t}$ and ego-motion $\hat{\bm{X}}^{t \to c}$~\cite{zhou2017unsupervised}. See Figure~\ref{fig:ssl} for an overview.
The general pixel-warping operation is defined as:
\begin{equation}%
    \hat{\bm{p}}^t = \pi \left({\hat{\bm{R}}}^{t \rightarrow c} \phi (\bm{p}^t, \hat{d}^t, \bm{i}) + \bm{\hat{t}}^{t \rightarrow c}, \bm{i}\right),
\end{equation}\label{eq:warp_mono}%
where $\bm{i}$ are camera intrinsic parameters modeling the geometry of the camera, which is required for both projection of 3D points $\bm{P}$ onto image pixels $\bm{p}$ via $\pi(\bm{P},\bm{i}) = \bm{p}$ and unprojection via $\phi(\bm{p}, \hat{d},\bm{i}) = \bm{P}$ assuming an estimated pixel depth of $\hat{d}$.
The camera parameters $\bm{i}$ are generally the standard pinhole model~\cite{hartley2000zisserman} defined by the $3 \times 3$ intrinsic matrix $\bm{K}$, but can include any differentiable model such as the Unified Camera Model family~\cite{usenko2018double} as described next.

\subsection{End-to-End Self-Calibration.}
\label{subsec:ucm}
\begin{figure}[!th]
    \centering
    \includegraphics[width=0.75\linewidth]{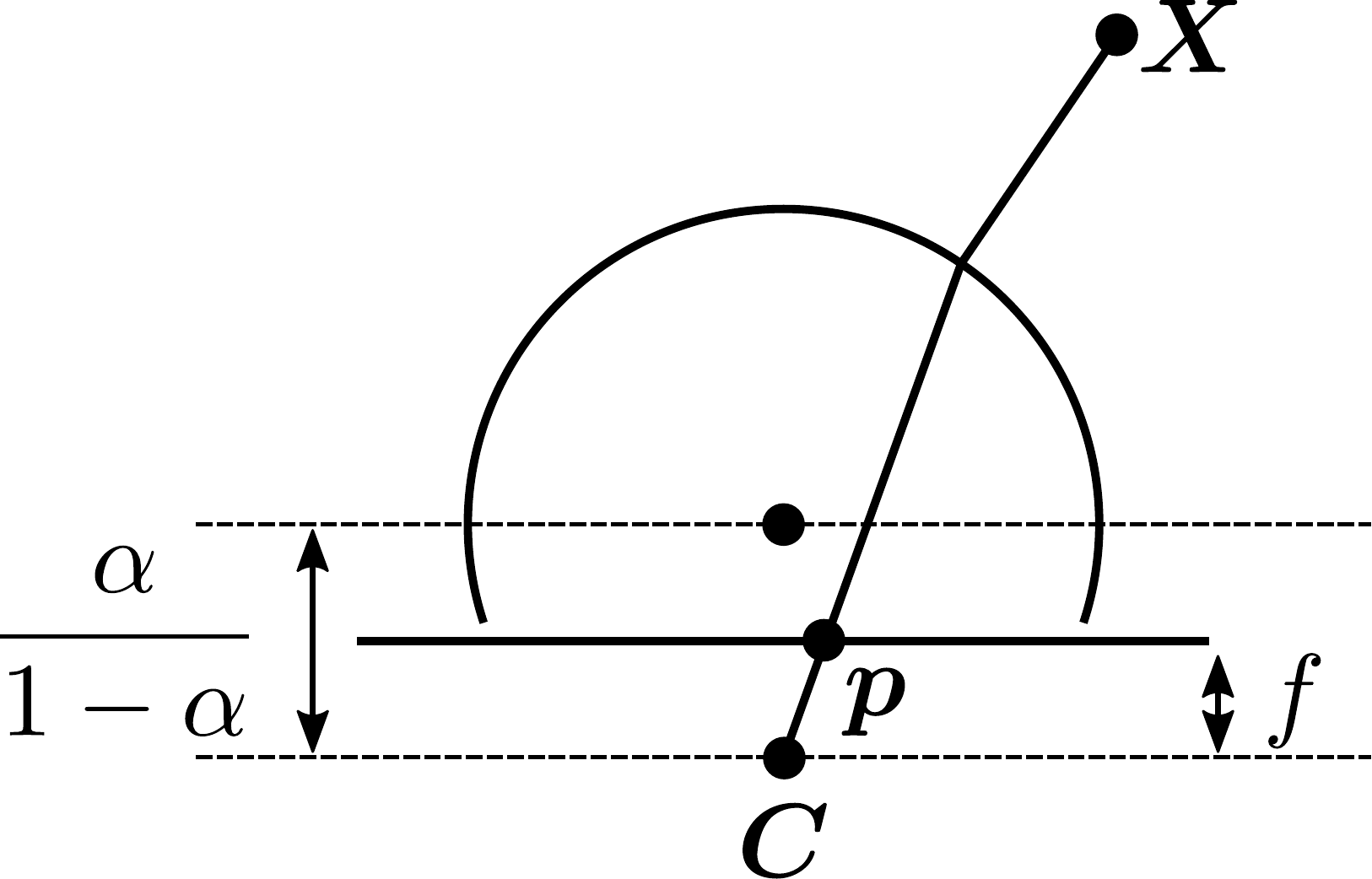}
    \caption{\textbf{The Unified Camera Model~\cite{usenko2018double} used in our self-calibration pipeline}. Points are projected onto a unit sphere before being projected onto an image plane of a standard pinhole camera offset by $\frac{\alpha}{1-\alpha}$ from the sphere center.}
    \label{fig:ucm_figure}
\vspace*{-2mm}
\end{figure}

UCM~\cite{geyer2000unifying} is a parametric global central camera model that uses only five parameters to represent a diverse set of camera geometries, including perspective, fisheye, and catadioptric. A 3D point is projected onto a unit sphere and then projected onto the image plane of a pinhole camera, shifted by $\frac{\alpha}{1-\alpha}$ from the center of the sphere (Fig.~\ref{fig:ucm_figure}). EUCM and DS are two extensions of the UCM model. EUCM replaces the unit sphere with an ellipse as the first projection surface, and DS replaces the one unit sphere with two unit spheres in the projection process.  We self-calibrate all three models (in addition to a pinhole baseline) in our experiments. For brevity, we only describe the original UCM and refer the reader to \citet{usenko2018double} for details on the EUCM and DS models.

There are multiple parameterizations for UCM \cite{geyer2000unifying}, and we use the one from~\citet{usenko2018double} since it has better numerical properties.  UCM extends the pinhole camera model $(f_x, f_y, c_x, c_y)$ with only one additional parameter $\alpha$. The 3D-to-2D projection of $\bm{P}=(x,y,z)$  is defined as

\begin{equation} \label{eq:ucm_proj}
    \bm{\pi}(\bm{P}, \bm{i}) = \begin{bmatrix}
f_x \frac{x}{\alpha d + (1-\alpha)z} 
\\ 
f_y \frac{y}{\alpha d + (1-\alpha)z} 
\end{bmatrix} +
\begin{bmatrix}
c_x \\
c_y
\end{bmatrix}
\end{equation}
where the camera parameters are $\bm{i} = (f_x, f_y, c_x, c_y, \alpha)$ and $d=\sqrt{x^2+y^2+z^2}$

The unprojection operation of pixel $\bm{p} = (u,v,1)$ at estimated depth $\hat{d}$ is:
\begin{equation}\label{eq:ucm_unproj}
    \phi(\bm{p}, \hat{d},\bm{i}) = \hat{d} \frac{\xi + \sqrt{1 + (1-\xi^2)r^2}}{1 + r^2}\begin{bmatrix} m_x \\ m_y \\ 1\end{bmatrix} - \begin{bmatrix} 0 \\ 0 \\ \hat{d} \zeta \end{bmatrix}
\end{equation}

where
\begin{subequations}
\begin{align}
    m_x &= \frac{u - c_x}{f_x}(1- \alpha)&
    m_y &= \frac{v - c_y}{f_y}(1- \alpha)&\\
    r^2 &= m_{x}^2 + m_{x}^2&
    \zeta &= \frac{\alpha}{1-\alpha}&
\end{align}
\end{subequations}

As shown in Equations~\ref{eq:ucm_proj} and \ref{eq:ucm_unproj}, the UCM camera model provides  closed-form projection and unprojection functions that are both differentiable.
Therefore, the overall architecture is end-to-end differentiable with respect to both neural network parameters (for pose and depth estimation) and camera parameters. This enables learning self-calibration end-to-end from the aforementioned view synthesis objective alone.
At the start of self-supervised depth and pose training, rather than pre-calibrating the camera parameters, we initialize the camera with ``default'' values based on image shape only (for a detailed discussion of the initialization procedure, please see Section~\ref{sec:perturbation_test}). 
Although the projection \eqref{eq:ucm_proj} and unprojection \eqref{eq:ucm_unproj} are initially inaccurate, they quickly converge to highly accurate camera parameters with sub-pixel re-projection error (see Table~\ref{table:reproj_error}). 

As we show in our experiments, our method combines flexibility with computational efficiency. Indeed, our approach enables learning from heterogeneous datasets with potentially vastly differing sensors for which separate parameters $\bm{i}$ are learned. As most of the parameters (in the depth and pose networks) are shared thanks to the decoupling of the projection model, this enables scaling up in-the-wild training of depth and pose networks. Furthermore, our method is efficient, with only one extra parameter relative to the pinhole model. This enables learning depth for highly-distorted catadioptric cameras at a much higher resolution than previous over-parametrized models ($1024 \times 1024$ vs. $384 \times 384$ for~\citet{vasiljevic2020neural}). 
Note that, in contrast to prior works~\cite{gordon2019depth, vasiljevic2020neural}, we learn intrinsics per-sequence rather than per-frame.
This increases stability compared to per-frame methods that exhibit frame-to-frame variability~\cite{vasiljevic2020neural}, and can be used over sequences of varying sizes.

\section{Experiments}
In this section we describe two sets of experimental validations for our architecture: (i) calibration, where we find that the re-projection error of our learned camera parameters compares favorably to target-based traditional calibration toolboxes; and (ii) depth evaluation, where we achieve state-of-the-art results on the challenging EuRoC MAV dataset.

\subsection{Datasets}

Self-supervised depth and ego-motion learning uses monocular sequences~\cite{zhou2017unsupervised, godard2019digging, gordon2019depth, packnet} or rectified stereo pairs~\cite{godard2019digging, superdepth} from forward-facing cameras~\cite{geiger2012we,packnet,caesar2020nuscenes}. Given that our goal is to learn camera calibration from raw videos in challenging settings, we use the standard KITTI dataset as a baseline, and focus on the more challenging and distorted EuRoC~\cite{burri2016euroc} fisheye sequences.

\noindent\textbf{{KITTI~\cite{geiger2012we}}}
We use this dataset to show that our self-calibration procedure is able to accurately recover pinhole intrinsics alongside depth and ego-motion. Following related work~\cite{zhou2017unsupervised, godard2019digging, gordon2019depth, packnet} we use the training protocol of~\cite{eigen2014depth}, including filtering static images as described by~\citet{zhou2017unsupervised}. The resulting training set contains of $39810$ images, with $697$ images left for evaluation. 

\noindent\textbf{{EuRoC~\cite{burri2016euroc}}} The dataset consists of a set of indoor MAV sequences with general six-DoF motion. Consistent with recent work~\cite{gordon2019depth}, we train using center-cropping and down-sample the images to a $384 \times 256$ resolution, while training and evaluating on the same split. For calibration evaluation, we follow~\citet{usenko2018double} and use the calibration sequences from the dataset. We evaluate the UCM, EUCM and DS camera models in terms of re-projection error.

\noindent\textbf{OmniCam~\cite{schonbein2014calibrating}} A challenging outdoor catadioptric sequence, containing 12000 frames captured by an autonomous car rig. As this dataset does not provide ground-truth depth information, we only provide qualitative results.

\subsection{Training Protocol}
We implement the group of unified camera models described in ~\cite{usenko2018double} as differentiable PyTorch~\cite{paszke2017automatic} operations, modifying the self-supervised depth and pose architecture of~\citet{godard2019digging} to jointly learn depth, pose, and the unified camera model intrinsics. We use a learning rate of $2$e-$4$ for the depth and pose network and $1$e-$3$ for the camera parameters.  We use a StepLR scheduler with $\gamma=0.5$ and a step size of $30$. All of the experiments are run for $50$ epochs. The images are augmented with random vertical and horizontal flip, as well as color jittering. We train our models on a Titan X GPU with 12\,GB of memory, with a batch size of $16$ when training on images with a resolution of $384 \times 256$. We note that our method requires significantly less memory than that of~\citet{vasiljevic2020neural} which learns a generalized camera model parameterized through a per-pixel ray surface. 

\captionsetup[table]{skip=6pt}

\begin{table}[t!]
\renewcommand{\arraystretch}{1.1}
\centering
{
\small
\setlength{\tabcolsep}{0.3em}
\begin{tabular}{lcc}
\toprule
\multirow{2}{*}{\bf{Method}} & \multicolumn{2}{c}{\emph{Mean Reprojection Error}} 
\\
\cmidrule{2-3}
  & 
\emph{Target-based} & 
\emph{Learned}
\\
\midrule
Pinhole & 1.950 & 2.230\\
UCM~\cite{geyer2000unifying}  &
0.145 & 0.249 \\
EUCM~\cite{khomutenko2015enhanced}  &
0.144 & 0.245 \\
DS~\cite{usenko2018double}  &
0.144  & 0.344 \\
\bottomrule
\end{tabular}
}
\caption{
\textbf{Mean reprojection error on EuRoC} at $256 \times 384$ resolution for UCM, EUCM and DS models using (left) AprilTag-based toolbox calibration Basalt ~\cite{usenko19nfr} and (right) our self-supervised learned (L) calibration.
Note that despite using no ground-truth calibration targets, our self-supervised procedure produces sub-pixel reprojection error.
}
\vspace{-3mm}
\label{table:reproj_error}
\end{table}

\subsection{Camera Self-Calibration}

We evaluate the results of the proposed self-calibration method on the EuRoC dataset; detailed depth estimation evaluations are provided in Sec.~\ref{subsec:results_depth}. To our knowledge, ours is the first direct calibration evaluation of self-supervised intrinsics learning; although~\citet{gordon2019depth} compare \textit{ground-truth} calibration to their per-frame model, they do not evaluate the re-projection error for their learned parameters. 

Following~\citet{usenko19nfr}, we evaluate our self-supervised calibration method on the family of unified camera models: UCM, EUCM, and DS, as well as the perspective (pinhole) model. As a lower bound, we use the Basalt~\cite{usenko19nfr} toolbox and compute camera calibration parameters for each unified camera model using the calibration sequences of the EuRoC dataset. We note that unlike Basalt, our method regresses the intrinsic calibration parameters directly from raw videos, without using any of the calibration sequences. 

\captionsetup[table]{skip=6pt}

\begin{table}[t!]
\renewcommand{\arraystretch}{1.1}
\centering
{
\small
\setlength{\tabcolsep}{0.3em}
\begin{tabular}{lcccccccc}
\toprule
\textbf{Method}  & 
$f_x$ &
$f_y$ &
$c_x$ &
$c_y$ & 
$\alpha$ &
$\beta$ &
$\xi$ &
$w$\\
\midrule
UCM (L) &
237.6 & 247.9 & 187.9 & 130.3 & 0.631 & \multirow{2}{*}{---} & \multirow{2}{*}{---}& \multirow{2}{*}{---}\\ 
UCM (B) & 235.4 & 245.1 & 186.5 & 132.6 & 0.650 \\ 
\midrule
EUCM (L) & 237.4 & 247.7 & 186.7 & 129.1 & 0.598 & 1.075 & \multirow{2}{*}{---} & \multirow{2}{*}{---}\\
EUCM (B) & 235.6 & 245.4 & 186.4 & 132.7 & 0.597 & 1.112\\
\midrule
DS (L) & 184.8 & 193.3 & 187.8 & 130.2 & 0.561 & \multirow{2}{*}{---} & -0.232 & \multirow{2}{*}{---}\\
DS (B) & 181.4 & 188.9 & 186.4 & 132.6 &  0.571 & & -0.230\\
\bottomrule
\end{tabular}
}
\caption{\textbf{Intrinsic calibration evaluation of different methods} on the EuRoC dataset, where B denotes intrinsics obtained from Basalt, and L denotes learned intrinsics.}
\label{table:intrinsic_numbers_compare}
\end{table}

\begin{figure}[!t]
  \centering
  \includegraphics[width=0.49\linewidth]{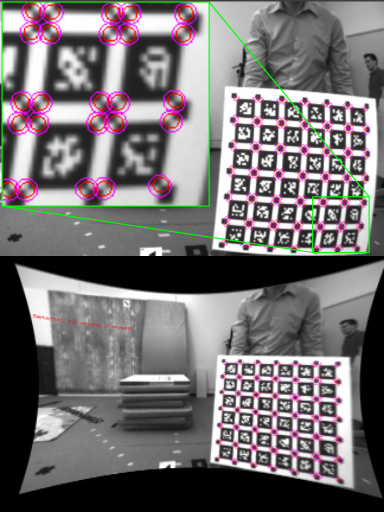}\hfil
  \includegraphics[width=0.49\linewidth]{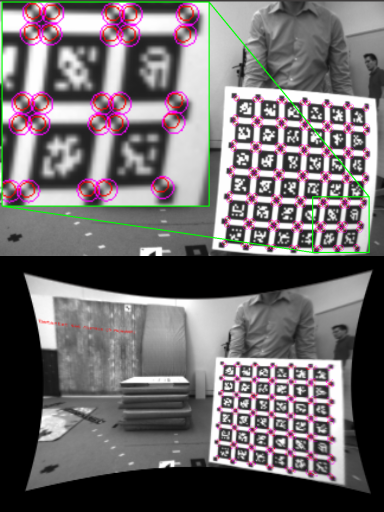}
\caption{\textbf{EuRoC rectification results} using images from the calibration sequences. Each column visualizes the results rendered using (left) the Basalt calibrated intrinsics and (right) our learned intrinsics. The top row shows that detected (small circles) and reprojected (big circles) corners are close using both calibration methods. The bottom row shows the same images after rectification.
}\label{fig:rectification}
\end{figure}
Table~\ref{table:reproj_error} summarizes our re-projection error results. We use the EuRoC AprilTag~\cite{olson2011apriltag} calibration sequences with Basalt to measure re-projection error using the full estimation procedure (Table~\ref{table:reproj_error} --- \textit{Target-based}) and learned intrinsics (Table~\ref{table:reproj_error} --- \textit{Learned}).  For consistency, we optimize for both intrinsics and camera poses for the baselines and only for the camera poses for the learned intrinsics evaluation. Note that with learned intrinsics, UCM, EUCM and DS models all achieve sub-pixel mean projection error despite the camera parameters having been learned from raw video data.

\captionsetup[table]{skip=6pt}

\begin{table}[t!]
\renewcommand{\arraystretch}{1.1}
\centering
{
\small
\setlength{\tabcolsep}{0.3em}
\begin{tabular}{lccccccc}
\toprule
\textbf{Perturbation}  & 
$f_x$ &
$f_y$ &
$c_x$ &
$c_y$  & 
$\alpha$ &
$\beta$ &
\textbf{MRE}

\\
\midrule

$I_{1.10}$ init & 242.3 & 253.6 & 189.5 & 130.7 & 0.5984 & 1.080 & 0.409\\
$I_{1.05}$ init & 241.3 & 252.3 & 188.5 & 130.5 & 0.5981 & 1.078 & 0.367\\
$I_c$ init & 240.2 & 251.4 & 187.9 & 130.0  & 0.5971 & 1.076 & 0.348\\
$I_{0.95}$ init & 239.5 & 250.9 & 187.8 & 129.2  & 0.5970 & 1.076 & 0.332\\
$I_{0.90}$ init & 238.8 & 249.6 & 187.7 & 129.1 & 0.5968 & 1.071 & 0.298\\

\midrule

$I_c$ & 235.6 & 245.4 & 186.4 & 132.7 & 0.597 & 1.112 & 0.144\\
\bottomrule
\end{tabular}
}
\caption{
\textbf{EUCM perturbation test results.} With perturbed initialization, all intrinsic parameters achieve sub-pixel convergence for mean reprojection error (\textbf{MRE}), with only a small offset to the Basalt calibration numbers.
}
\label{table:perturbation}
\end{table}

\begin{figure}[!t]
  \centering
    \subfloat[$f_x$]{\includegraphics[width=0.23\textwidth]{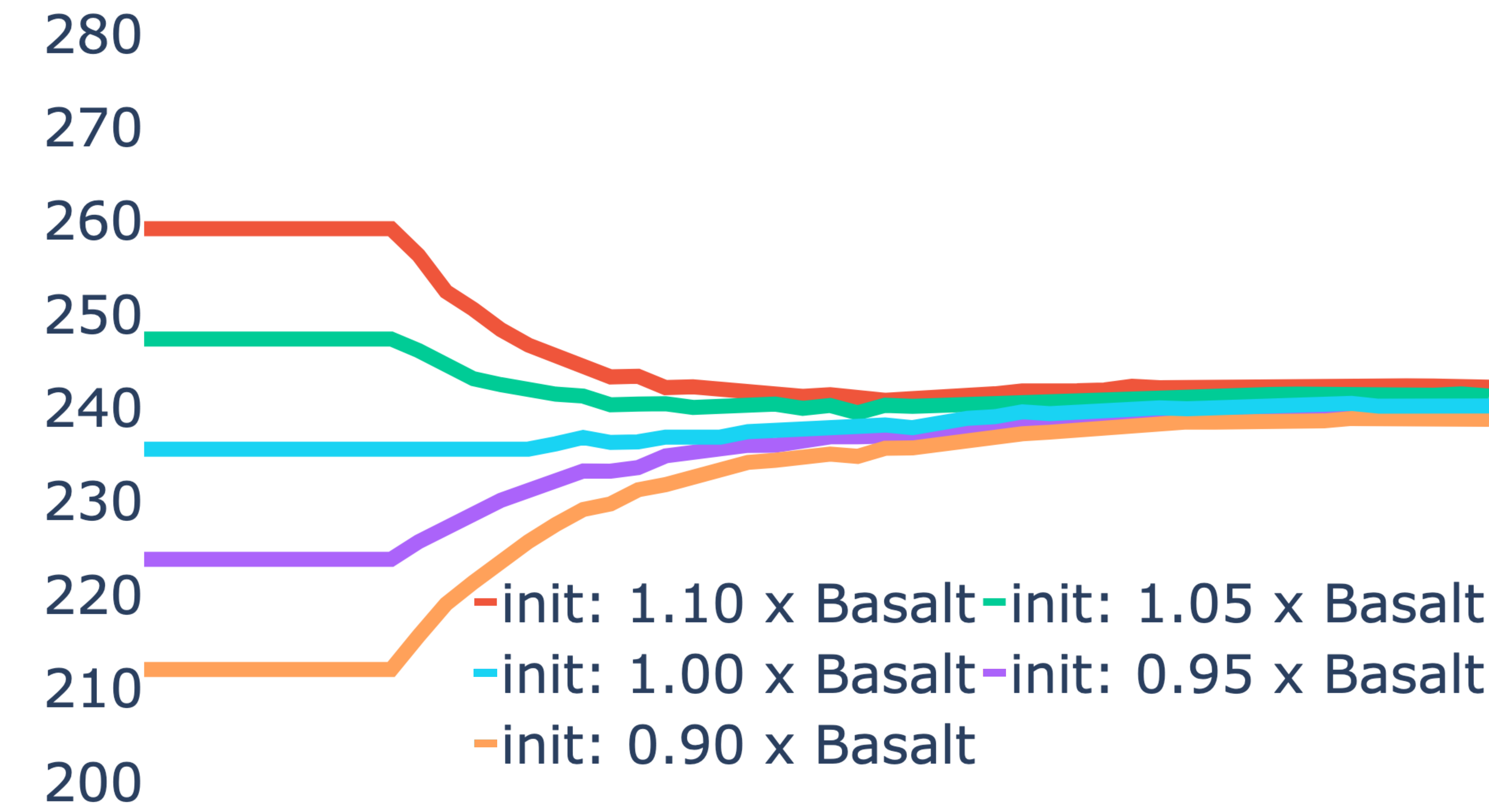}}\hfil
    \subfloat[$f_y$]{\includegraphics[width=0.23\textwidth]{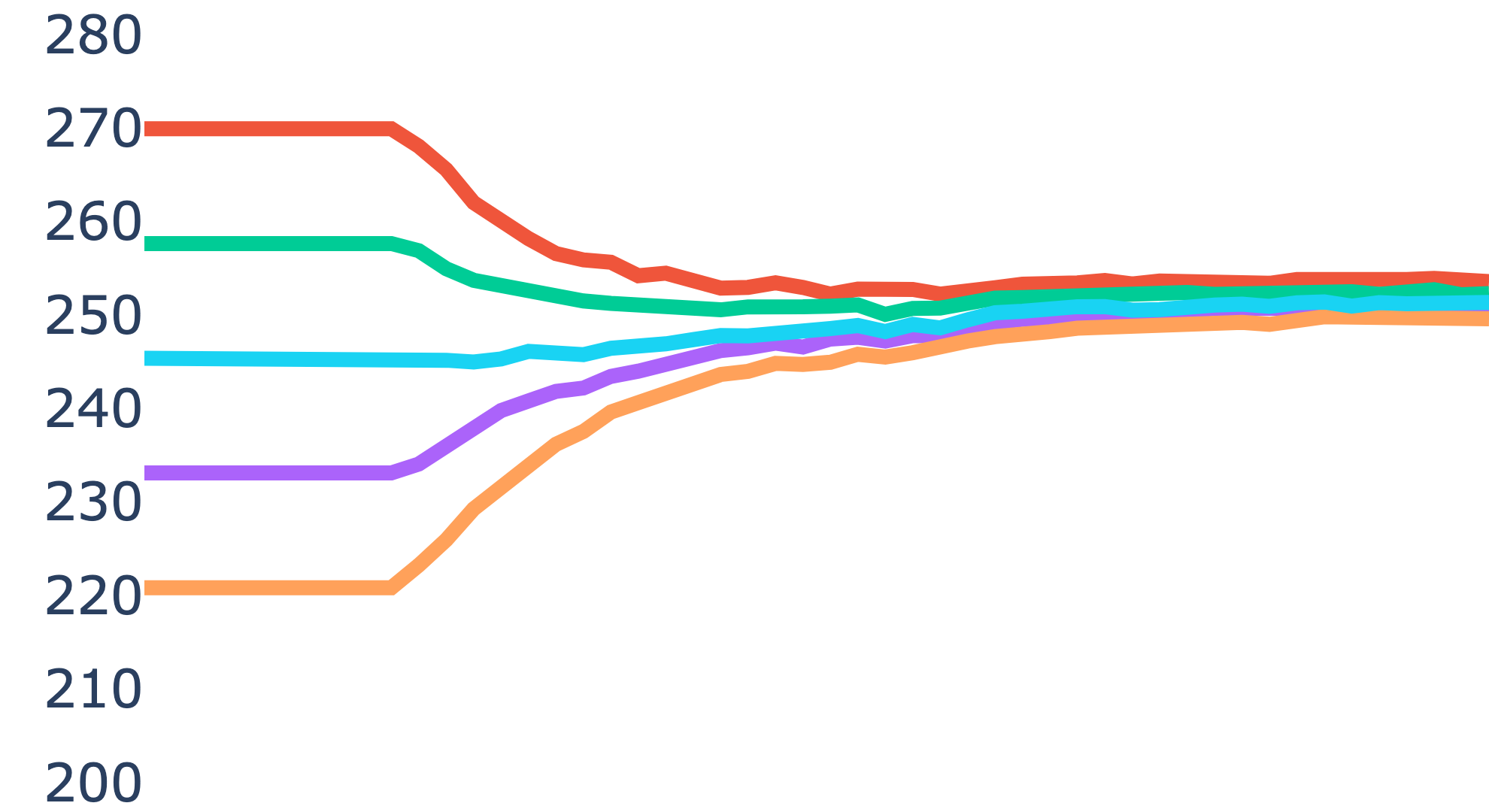}}\\
    \subfloat[$c_x$]{\includegraphics[width=0.23\textwidth]{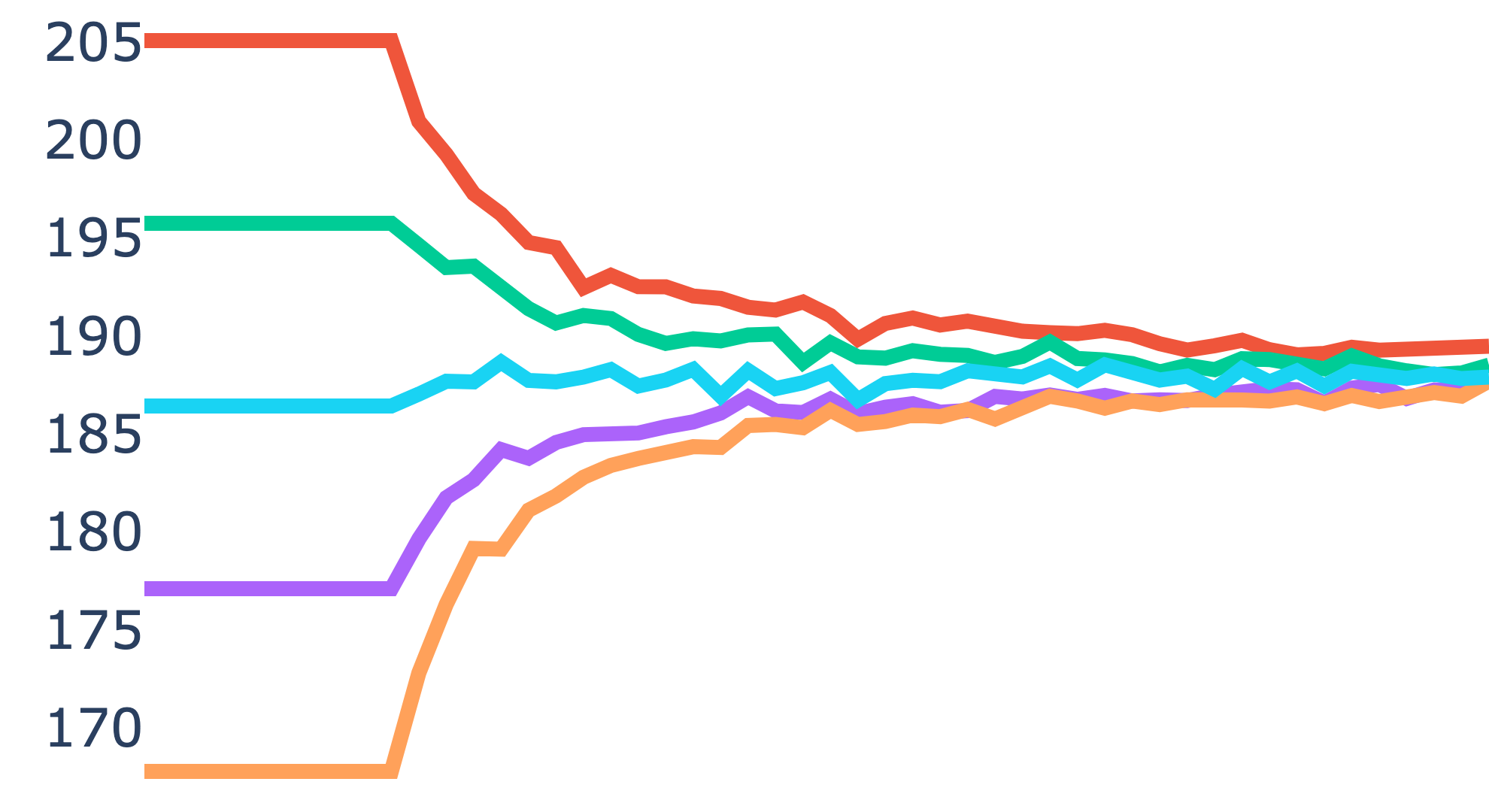}}\hfil
    \subfloat[$c_y$]{\includegraphics[width=0.23\textwidth]{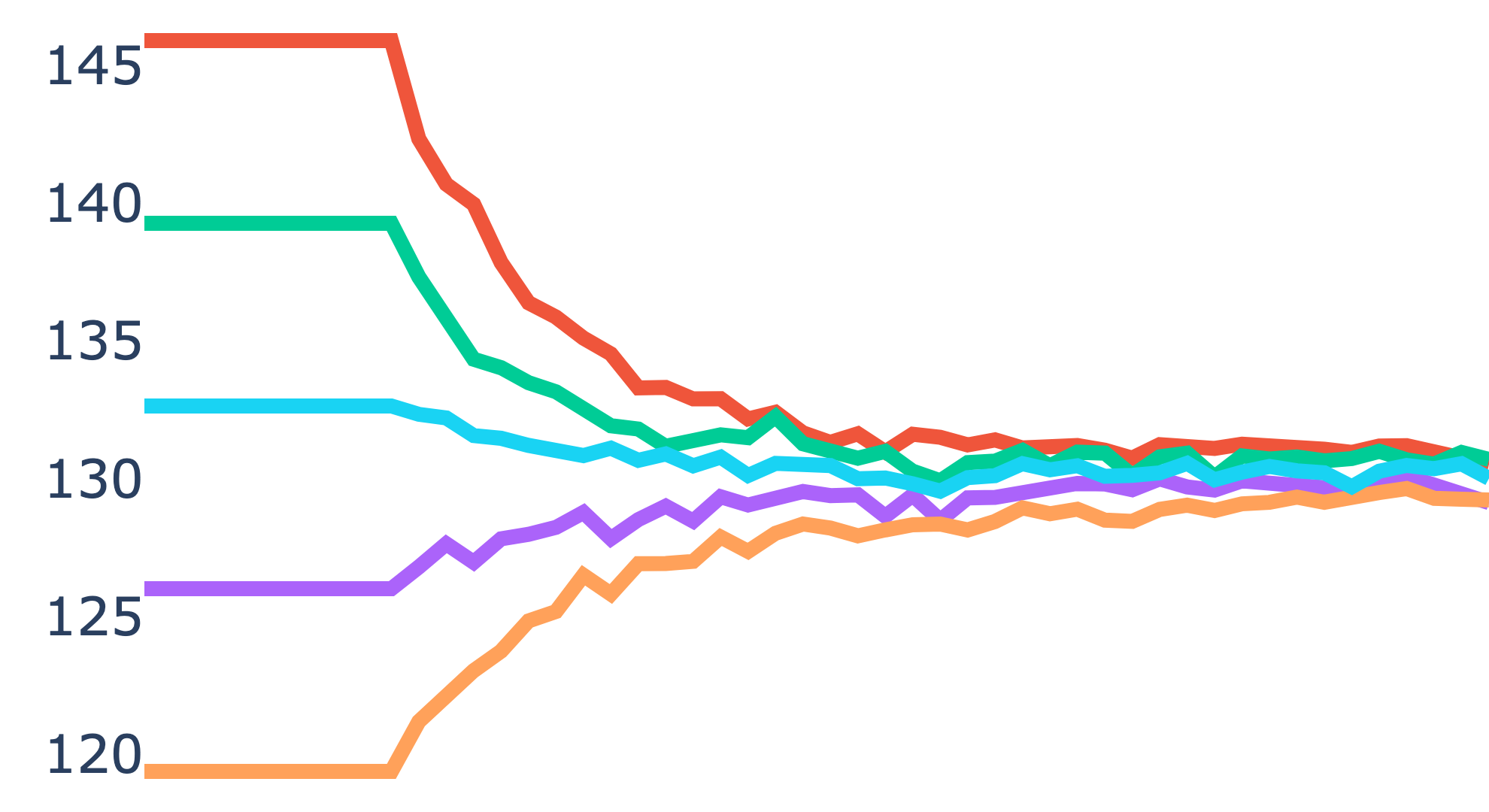}}\\
    \subfloat[$\alpha$]{\includegraphics[width=0.23\textwidth]{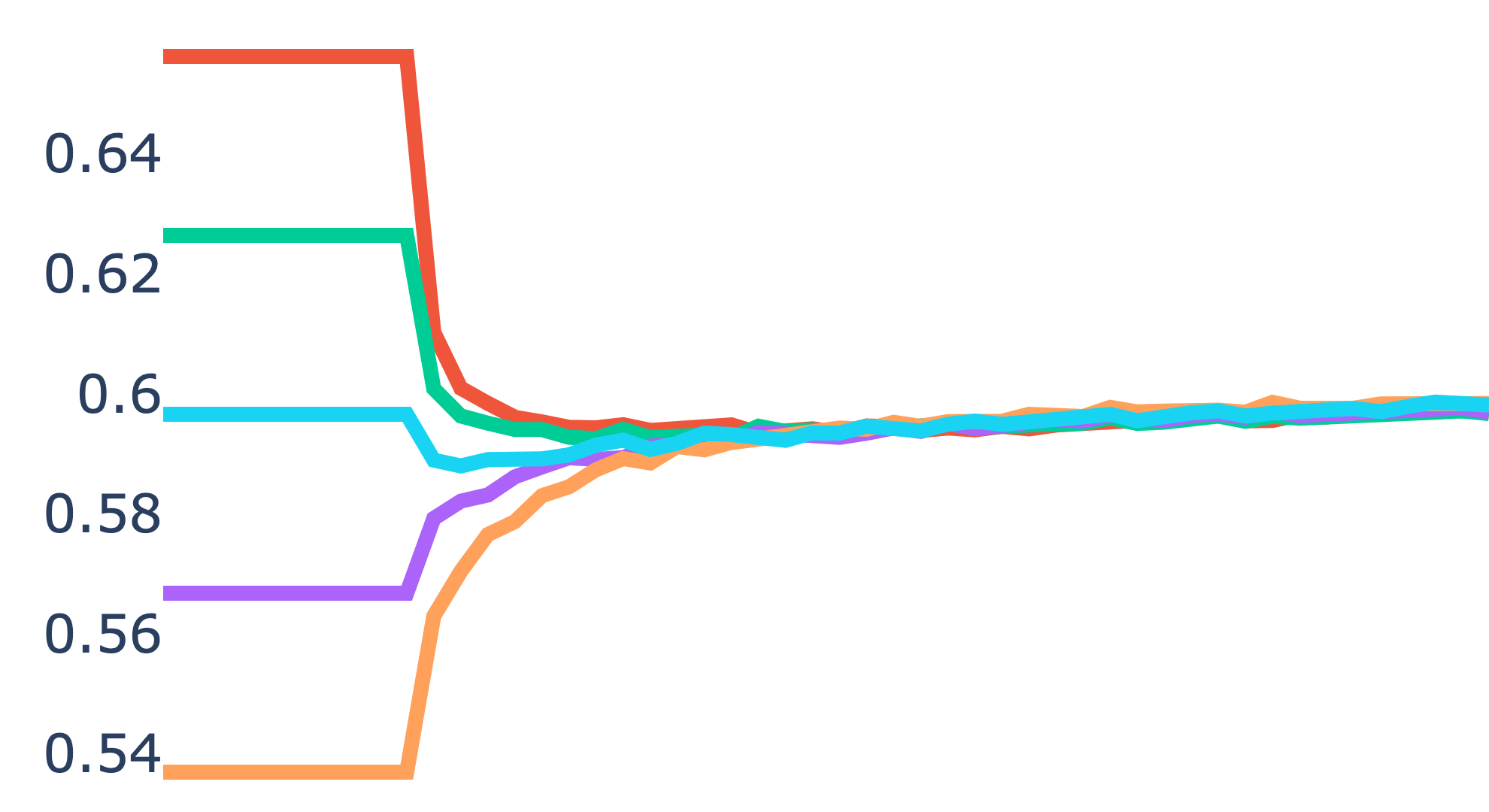}}\hfil
    \subfloat[$\beta$]{\includegraphics[width=0.23\textwidth]{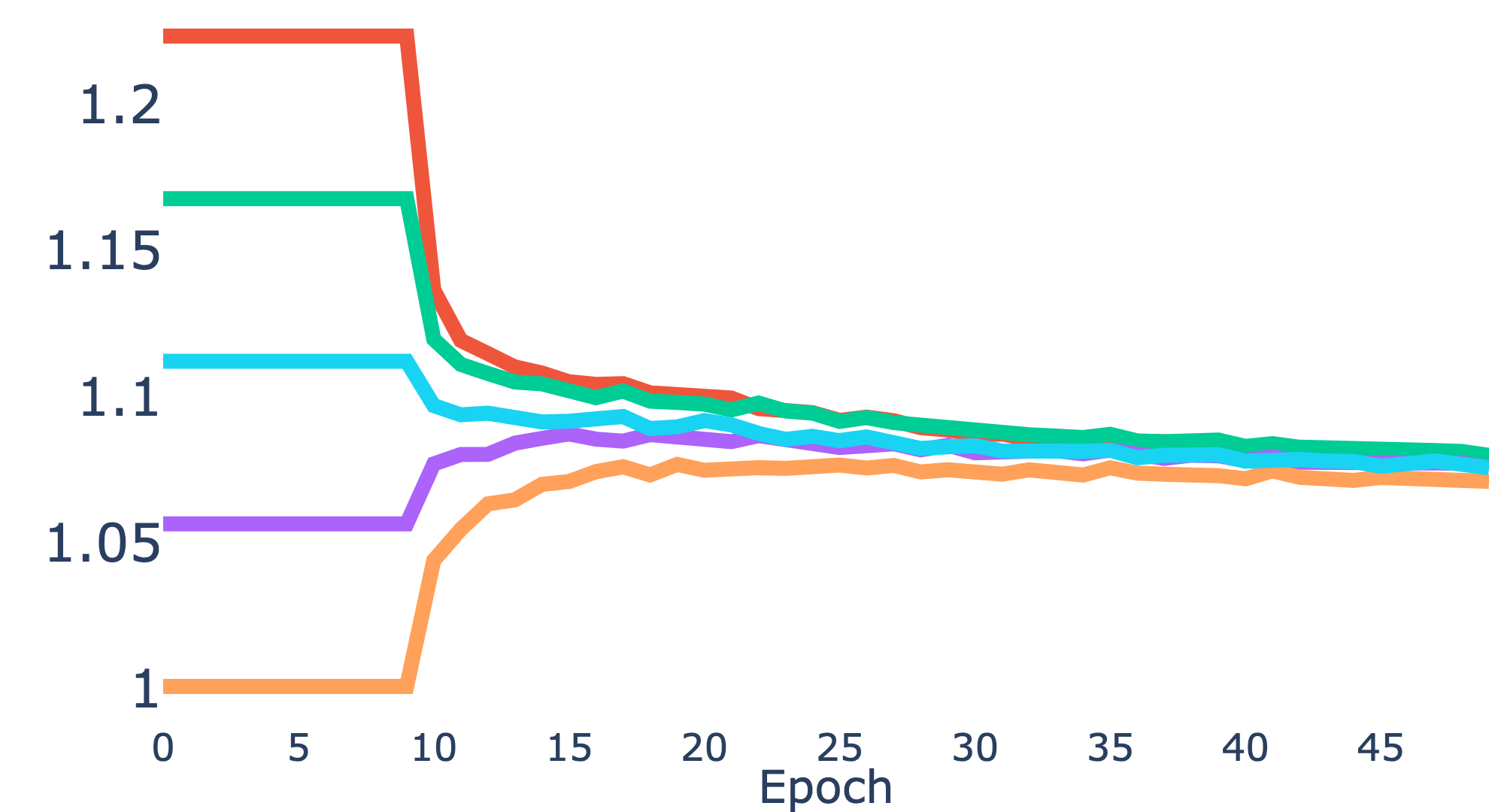}}
  \caption{\textbf{EuRoC perturbation test}, showing how our proposed learning-based method is able to recover from changes in camera parameters for online self-calibration.}

  \label{fig:perturbation}
  \vspace{-3mm}
\end{figure}

Table~\ref{table:intrinsic_numbers_compare} compares the target-based calibrated parameters to our learned parameters for different camera models trained on the \textit{cam0} sequences of the EuRoC dataset. Though the parameter vectors were initialized with no prior knowledge of the camera model and updated purely based on gradients from the reprojection error, they converge to values very close to the output of a procedure that uses bundle adjustment on calibrated image sequences.

\subsection{Camera Rectification}
Using our learned camera parameters, we rectify calibration sequences on the EuRoC dataset to demonstrate the quality of the calibration.  EuRoC was captured with a fisheye camera and exhibits a high degree of radial distortion that causes the straight edges of the checkerboard grid to be curved.  In Figure~\ref{fig:rectification}, we see that our learned parameters allow for the rectified grid to track closely to the true underlying checkerboard.

\subsection{Camera Re-calibration: Perturbation Experiments} \label{sec:perturbation_test}
Thus far, we have assumed to have no prior knowledge of the camera calibration. In many real-world robotics settings, however, one may want to re-calibrate a camera based on a potentially incorrect prior calibration.  Generally, this requires the capture of new calibration data.  Instead, we can initialize our parameter vectors with this initial calibration (in this setting, a perturbation of Basalt calibration of the EUCM model) and see the extent to which self-supervision can nudge the parameters back to their ``true value''.   

Given Basalt parameters $I_c=[f_x, f_y, c_x, c_y, \alpha, \beta]$, we perturb them as $I_{1.1}=1.1\times I_c$, $I_{1.05}=1.05\times I_c$, $I_{0.95}=0.95\times I_c$, $I_{0.9}=0.9\times I_c$ and initialize the camera parameters at the beginning of training with these values. All runs have warm start, i.e., freezing the gradients for the intrinsics for the first $10$ epochs while we train the depth and pose networks.  As Figure~\ref{fig:perturbation} shows, our method converges to within $3\%$ of the Basalt estimate for each parameter.  Table~\ref{table:perturbation} provides the values of the converged parameters along with the mean projection error (MRE) for each experiment.

\captionsetup[table]{skip=6pt}

\begin{table}[t!]
\renewcommand{\arraystretch}{1.1}
\centering
{
\small
\setlength{\tabcolsep}{0.2em}
\begin{tabular}{llcccc}
\toprule
\textbf{Method}  &  Camera &
Abs Rel$\downarrow$ &
Sq Rel$\downarrow$ &
RMSE$\downarrow$ &
$\delta_{1.25}$ $\uparrow$ 
\\
\toprule
\citet{gordon2019depth} &
K &
0.129 & 0.982 & 5.23 & 0.840\\ 
\citet{gordon2019depth} & 
L(P) &
0.128 & 0.959 & 5.23 & 0.845 \\ 
\citet{vasiljevic2020neural} &
K(NRS) &
0.137 & 0.987 & 5.33 & 0.830 \\ 
\citet{vasiljevic2020neural} &
L(NRS) &
0.134 & 0.952 & 5.26 & 0.832 \\ 
\midrule
Ours & 
L(P) &
0.129 & \textbf{0.893} & 4.96 & 0.846 \\ 
Ours &
L(UCM) &
\textbf{0.126} & 0.951 & \textbf{4.89} &  \textbf{0.858} \\ 
\bottomrule
\end{tabular}
}
\caption{
\textbf{Quantitative depth evaluation on the KITTI \cite{burri2016euroc} dataset}, using the standard \emph{Eigen} split and the \emph{Garg} crop, for distances up to 80m (with median scaling). K and L($\cdot$) denote known and learned intrinsics, respectively.
 P means pinhole model.}
\label{table:kitti_depth}
\end{table}
\captionsetup[table]{skip=6pt}

\begin{table}[t!]
\renewcommand{\arraystretch}{1.1}
\centering
{
\small
\setlength{\tabcolsep}{0.3em}
\begin{tabular}{lccccc}
\toprule
\textbf{Method}  & 
Camera  & 
Abs Rel$\downarrow$ &
Sq Rel$\downarrow$ &
RMSE$\downarrow$ &
$\alpha_{1}$ $\uparrow$ 
\\
\midrule
\citet{gordon2019depth} & 
PB 
&
0.332 & 0.389 & 0.971 & 0.420 \\ 
\citet{vasiljevic2020neural}
&
NRS 
&
0.303 & 0.056 & 0.154 & 0.556 \\ 
\midrule

Ours & UCM & 0.282 & 0.048 & 0.141 & 0.591 \\ 
Ours &EUCM & \textbf{0.278} & \textbf{0.047} & \textbf{0.135} & \textbf{0.598} \\ 
Ours &DS & \textbf{0.278} & 0.049 & 0.141 & 0.584 \\ 

\bottomrule
\end{tabular}
}
\caption{
\textbf{Quantitative depth evaluation of different methods on the EuROC \cite{burri2016euroc} dataset}, using the evaluation procedure in~\cite{gordon2019depth} with center cropping.  The training data consists of ``Machine Room'' sequences and the evaluation is on the ''Vicon Room 201'' sequence (with median scaling). PN means plum-bob model.}
\label{table:euroc_depth_MH}
\end{table}

\subsection{Depth Estimation}
\label{subsec:results_depth}
\begin{figure}[!t]
 \centering
 \includegraphics[width=1.0\linewidth]{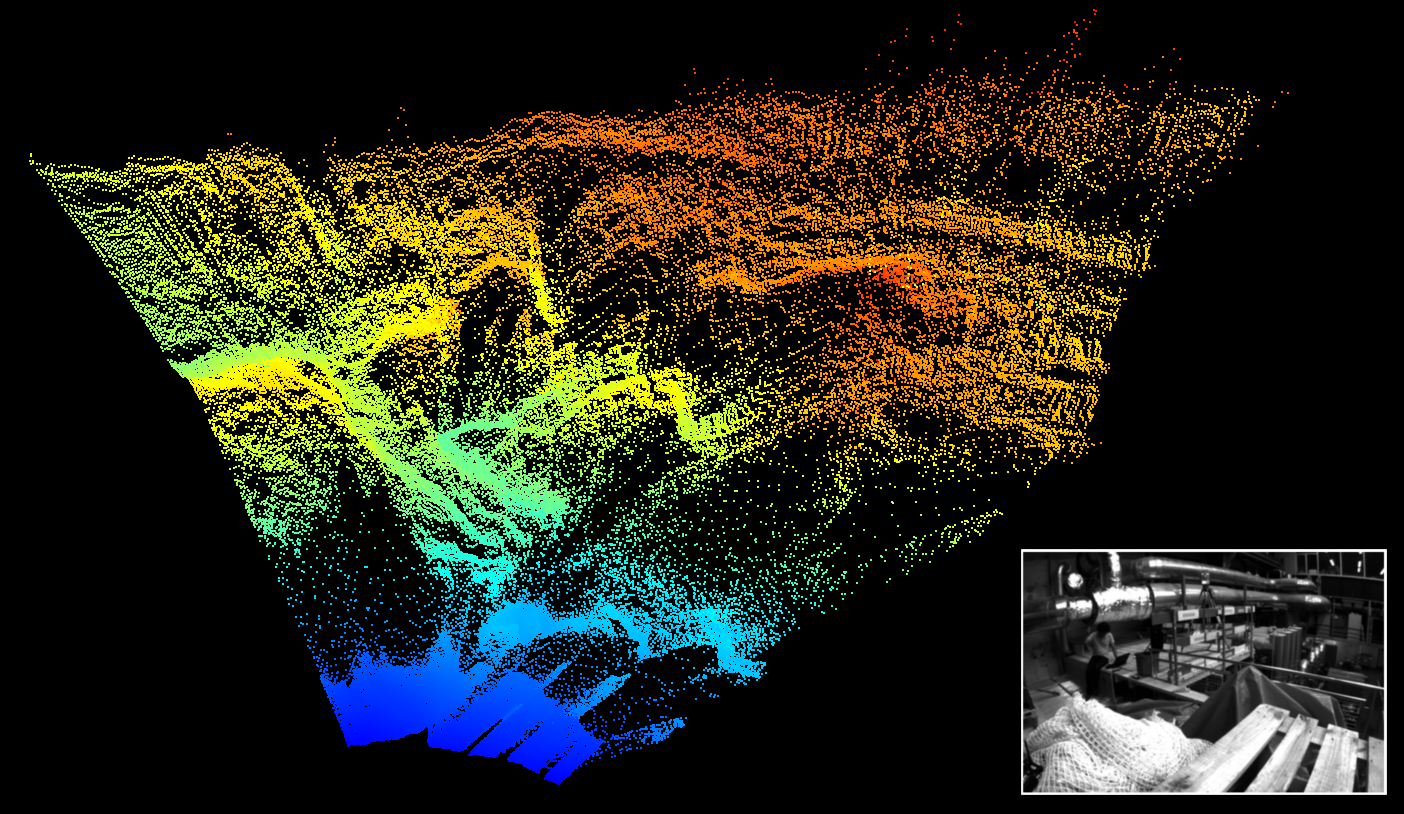}
\caption{\textbf{Self-supervised monocular pointcloud} for EuRoC, obtained by unprojecting predicted depth with our learned camera parameters (input image on the bottom right).} \label{fig:pointcloud_euroc}
\vspace{-3mm}
\end{figure}

\begin{figure}[!t]
  \centering
 \subfloat[EuRoC]{
  \includegraphics[width=0.32\linewidth]{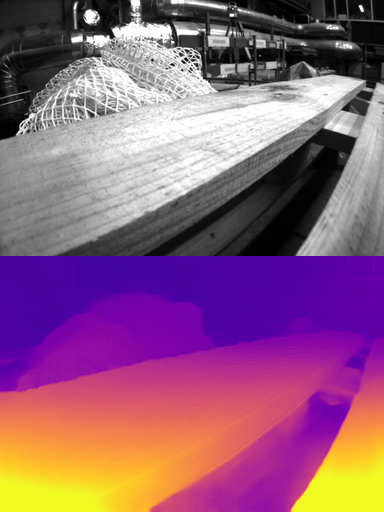}
  \includegraphics[width=0.32\linewidth]{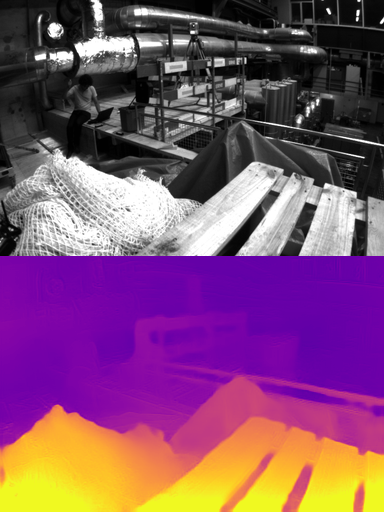}
  \includegraphics[width=0.32\linewidth]{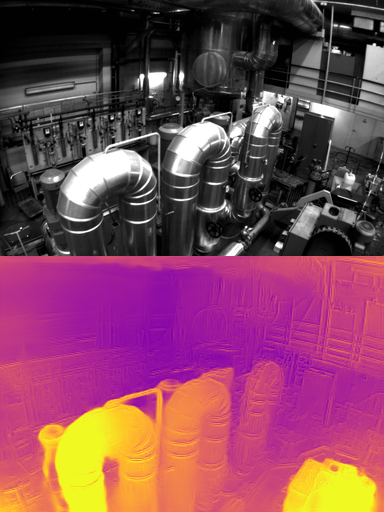}
}\label{fig:depth_euroc}
\subfloat[OmniCam]{
  \includegraphics[width=0.32\linewidth]{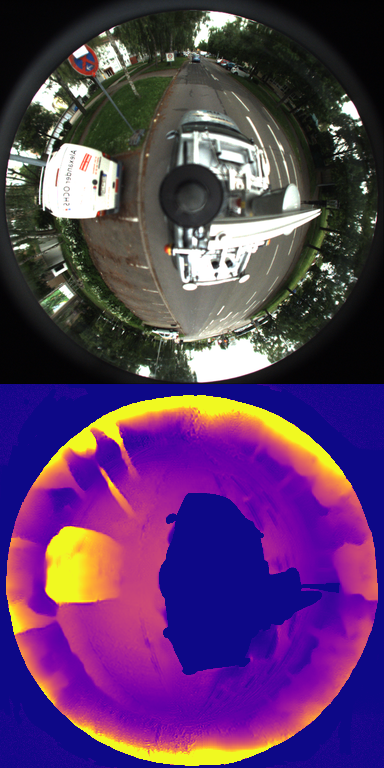}
  \includegraphics[width=0.32\linewidth]{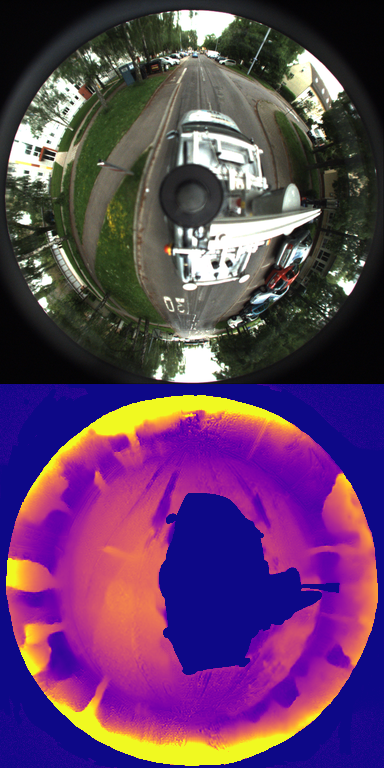}
  \includegraphics[width=0.32\linewidth]{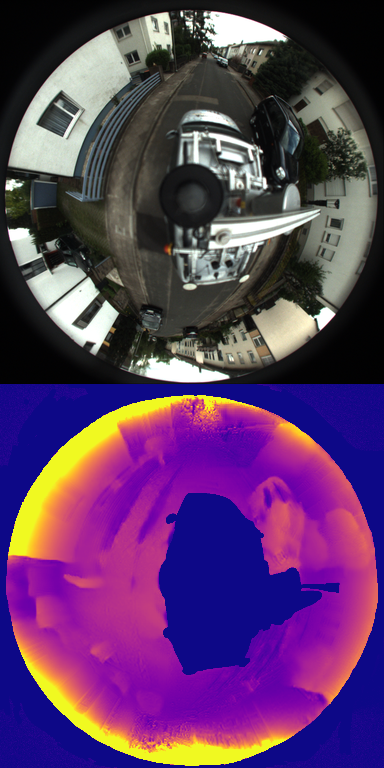}
 }
  \caption{\textbf{Qualitative depth estimation results} on non-pinhole datasets with (a) fisheye and (b) catadioptric images.}
  \label{fig:depth_omnicam}
\vspace{-3mm}
\end{figure}

While we use depth and pose estimation as proxy tasks for camera self-calibration, the unified camera model framework allows us to achieve meaningful results compared to prior camera-learning-based approaches (see Figures~\ref{fig:pointcloud_euroc} and \ref{fig:depth_omnicam}). 

\noindent\textbf{KITTI results.} Table~\ref{table:kitti_depth} presents the results of our method on the KITTI dataset. We note that our approach is able to model the simple pinhole setting, achieving results that are on par with approaches that are  tailored specifically to this camera geometry. Interestingly, we see an increase in performance using the UCM model, which we attribute to the ability to further account for and correct calibration errors.

\noindent\textbf{EuRoC results.}
Compared to KITTI, EuRoC is a significantly more challenging dataset that involves cluttered indoor sequences with six-DoF motion. Compared to the per-frame distorted camera models of \citet{gordon2019depth} and \citet{vasiljevic2020neural}, we achieve significantly better absolute relative error, especially with EUCM, where the error is reduced by $16\%$ (see Table~\ref{table:euroc_depth_MH}). We also train NRS~\cite{vasiljevic2020neural} on this dataset for further comparison, using the official repository.  

\captionsetup[table]{skip=6pt}

\begin{table}[t!]
\renewcommand{\arraystretch}{1.1}
\centering
{
\small
\setlength{\tabcolsep}{0.3em}
\begin{tabular}{lcccccc}
\toprule
\textbf{Dataset}  & 
Abs Rel$\downarrow$ &
Sq Rel$\downarrow$ &
RMSE$\downarrow$ &
$\alpha_{1}$ $\uparrow$ & 
$\alpha_{2}$ $\uparrow$ &
$\alpha_{3}$ $\uparrow$
\\
\midrule
EuRoC~\cite{gordon2019depth}  &
0.265 & \textbf{0.042} & 0.130 & 0.600 & 0.882 & \textbf{0.966} \\
EuRoC+KITTI &
\textbf{0.244} & 0.044 & \textbf{0.117} & \textbf{0.742} & \textbf{0.907} & 0.961 \\

\bottomrule
\end{tabular}
}
\caption{
\textbf{Quantitative multi-dataset depth evaluation} on EuRoC (without cropping and with median scaling). 
}
\label{table:multi_depth}
\end{table}

\noindent\textbf{Combining heterogeneous datasets.}
One of the strengths of the unified camera model is that it can represent a wide variety of cameras without prior knowledge of their specific geometry. As long as we know which sequences come from which camera, we can learn separate calibration vectors that share the same depth and pose networks. This is particularly useful as a way to improve performance on smaller datasets, since it enables one to take advantage of unlabeled data from other sources. To evaluate this property, we experimented with mixing KITTI and EuRoC.  In this experiment, we reshaped the KITTI images to match those in the EuRoC dataset (i.e.,  $384 \times 256$). As Table~\ref{table:multi_depth} shows, our algorithm is able to take advantage of the KITTI images to improve performance on the EuRoC depth evaluation.

\subsection{Computational Cost}
Our work is closely related to the learned general camera model (NRS) of~\citet{vasiljevic2020neural} given that in both works the parameters of a central general camera model are learned in a self-supervised way.  Being a per-pixel model, NRS is more general than ours and can handle settings where there is local distortion, which a global camera necessarily cannot model.  However, the computational requirements of the per-pixel NRS are significantly higher. For example, we train on EuRoC images with a resolution of $384 \times 256$ with a batch size of $16$, which consumes about $6$\,GB of GPU memory.  Each epoch takes about $15$ minutes.

On the same GPU, NRS uses $16$\,GB of GPU memory with a batch size of $1$ to train on the same sequences, running one epoch in about $120$ minutes. This is due to the high-dimensional (yet approximate) projection operation required for a generalized camera. Thus, we trade some degree of generality for significantly higher efficiency than prior work, with higher accuracy on the EuRoC dataset (see Table~\ref{table:euroc_depth_MH}).

\section{Conclusion}

We proposed a procedure to self-calibrate a family of general camera models using self-supervised depth and pose estimation as a proxy task.  We rigorously evaluated the quality of the resulting camera models, demonstrating sub-pixel calibration accuracy comparable to manual target-based toolbox calibration approaches. Our approach generates per-sequence camera parameters, and can be integrated into any learning procedure where calibration is needed and the projection and un-projection operations are interpretable and differentiable. As shown in our experiments, our approach is particularly amenable to online re-calibration, and can be used to combine datasets of different sources, learning independent calibration parameters while sharing the same depth and pose network. 

\clearpage
{\small
\bibliographystyle{IEEEtranN}
\bibliography{references}
}



\end{document}